\documentclass[runningheads]{llncs}
\usepackage{graphicx}
\usepackage{amsmath}
\usepackage{xcolor}
\usepackage{multirow}
\usepackage{algorithm, algorithmic}
\usepackage{pifont}
\newcommand{\checkmark}{\text{\ding{51}}}
\newcommand{\xmark}{\text{\ding{55}}}
\begin{document}
\title{Federated Adversarial Learning for Robust Autonomous Landing Runway Detection}
%
%
\author{Yi Li, Plamen Angelov, Zhengxin Yu, Alvaro Lopez Pellicer, Neeraj Suri}

%
%
\institute{Computing and Communications, Lancaster University, UK}
\maketitle              
\begin{abstract}
As the development of deep learning techniques in autonomous landing systems continues to grow, one of the major challenges is trust and security in the face of possible adversarial attacks. In this paper, we propose a federated adversarial learning-based framework to detect landing runways using paired data comprising of clean local data and its adversarial version. Firstly, the local model is pre-trained on a large-scale lane detection dataset. Then, instead of exploiting large instance-adaptive models, we resort to a parameter-efficient fine-tuning method known as scale and shift deep features (SSF), upon the pre-trained model. Secondly, in each SSF layer, distributions of clean local data and its adversarial version are disentangled for accurate statistics estimation. To the best of our knowledge, this marks the first instance of federated learning work that address the adversarial sample problem in landing runway detection. Our experimental evaluations over both synthesis and real images of Landing Approach Runway Detection (LARD) dataset consistently demonstrate good performance of the proposed federated adversarial learning and robust to adversarial attacks.
\end{abstract}
\vspace{-2em}
\section{Introduction}
An unmanned aerial vehicle (UAV), commonly known as a drone, refers to an aircraft operated without a human pilot on board \cite{uav}. Drones find applications in various fields, such as photography, research, surveillance, defense, and space exploration. Enhancing the autonomy of aircraft is crucial as it reduces the cognitive load on pilots, ensuring safety in civil aviation \cite{uav2}. Despite these advancements, unmanned aerial vehicles face challenges during the approach and landing phases. Recent progress in computer vision and embedded hardware platforms has positioned vision-based algorithms as an efficient direction for guiding and navigating during the landing stage. A vision-based landing system must be capable of detecting runways, from a distance to close proximity, in high-resolution images. Autonomous Landing Runway Detection (ALRD) aims to identify and determine a suitable runway for an aircraft to land at an airport \cite{lrd}. 


While there is considerable practical and commercial interest in autonomous landing systems within the aerospace field, there is currently a shortage of open-source datasets containing aerial images. To tackle this gap, a recent introduction of the Landing Approach Runway Detection (LARD) dataset \cite{lard} aims to provide a collection of high-quality aerial images specifically designed for the task of runway detection during approach and landing phases. The dataset primarily comprises images generated using conventional landing trajectories, where the possible positions and orientation of the aircraft during landing are defined within a generic landing approach cone. 




In recent times, significant advancements in deep learning techniques have greatly enhanced their application in vision-based ALRD, resulting in commendable performance gains \cite{lrd1}\cite{lrd}\cite{lrd2}. Typically, a neural network is trained using an extensive dataset of vision-based landing data to predict runways at varying distances in images captured by UAV cameras. However, the vulnerability to adversarial attacks in deep learning techniques poses a considerable risk in real-world ALRD applications. Adversarial attacks in federated learning, especially at the local image level, can manifest as data poisoning, data tampering, and privacy attacks. External attackers may inject poisoned data into local data \cite{ssci}. This may involve introducing adversarial examples crafted to deceive the local model during training. Conversely, in traditional large-scale neural network-based federated learning \cite{largefl}\cite{largefl1}, a computational cost issue arises where each client is required to train an individual model. However, as clients share the same task and data, they can potentially leverage shared features and the majority of weights.

The contributions of this paper are summarized as follows:

$\bullet $  We propose an approach based on federated learning for ALRD against adversarial attacks. Firstly, we pre-train a neural network on a large-scale lane detection dataset \cite{tvtLANE}. Next, the network is fine-tuned with paired images, \textit{i.e.}, clean and adversarial images in $M$ clients. 

$\bullet $ Differing from conventional large-scale model-based federated learning \cite{largefl}\cite{largefl1}, the weights of the proposed pre-trained model are frozen. In each client, we scale and shift the deep features (SSF) to leverage the representation abilities of large-scale pre-training models to achieve good performance on downstream tasks by fine-tuning a few trainable parameters. 

$\bullet $  Distributions of clean local data and its adversarial version are disentangled for accurate statistics estimation. Consequently, deep features of paired images are jointly learned at each layer of the local model for the model to learn downstream information from adversarial images. 


\vspace{-1em}
\section{Related Works} \label{sec:relatedworks}
\vspace{-1em}
\subsection{Deep Learning for ALRD}
The development of methods for detecting landing runways is pivotal for ensuring the security of autonomous aerial systems. These methods can be broadly categorized into two main approaches based on the data type: conventional image processing and deep learning-based image processing, video processing, multi-sensor fusion, and end-to-end learning. Firstly, conventional image processing involves techniques that manipulate and analyze images using traditional, rule-based algorithms such as edge detection \cite{edge}, contour analysis \cite{cont}, and color-based segmentation \cite{color}. Secondly, deep learning-based techniques \cite{lrd1}\cite{lrd2} have been developed and applied in image processing to improve the accuracy of aerial image detection. Finally, in contrast to conventional feature learning methods, some recent deep learning techniques \cite{lrd6} learn a mapping from raw sensor inputs to runway detection without explicit feature engineering. While these methods achieve high runway detection accuracy over image- or video-based datasets, they face three significant challenges \cite{lrd}: 1) variable groundtruth sizes over parameters, \textit{e.g.}, along track distance and vertical path angle, 2) robustness to adversarial attacks, 3) low execution time and computational costs.
\vspace{-1em}
\subsection{Adversarial Attacks}
Recent studies have highlighted the susceptibility of trained neural networks to compromise through adversarial samples, even with imperceptible perturbations that evade human detection \cite{attacks}. This raises significant safety concerns regarding the deployment of such networks in real-world applications, including critical domains like autonomous driving and clinical settings \cite{adver}.

The threat model categorizes existing adversarial attacks into three types: white-box, gray-box, and black-box attacks, differing in the level of knowledge possessed by adversaries \cite{IJCNN}. Within these threat models, various attack algorithms for generating adversarial samples have been proposed, including the Fast Gradient Sign Method (FGSM) \cite{fgsm}, Projected Gradient Descent (PGD) \cite{pgd}, Semantic Similarity Attack on High-Frequency Components (SSAH) \cite{SSAH}, Carlini $\&$ Wagner (CW) \cite{cw}, DeepFool \cite{deepfool}\cite{cvprw1}, Basic Iterative Method (BIM) \cite{bim}, and Jacobian-based Saliency Map Attack \cite{jsma}.
\vspace{-1em}
\subsection{Federated Learning}
Federated learning is a distributed machine learning approach that enables multiple clients to train a model collaboratively by using their local data without sharing \cite{fl0}. A key characteristic of federated learning is that the training process takes place locally on each device. Instead of sharing raw data, only model updates are exchanged among the devices throughout the training process \cite{fl2}. This mechanism effectively reduce the risk of data exposure, making it a privacy-preserving approach for collaborative model training over devices \cite{fl1}. 

Most existing federated learning methods achieve promising performance over a wide range of tasks. However, these methods have two limitations. Firstly, \cite{flat}\cite{flat1}\cite{flat2} have shown that local data can be vulnerable to adversarial attacks. If an adversary can inject malicious data during the training phase, they may subtly alter the model's behavior. This could lead to vulnerabilities that can be exploited during inference. Secondly, the number of parameters of pre-trained models is usually very large \cite{large}\cite{large1}, and simply fine-tuning the full model undoubtedly yields a huge amount of communication cost in federated learning algorithms. These limitations are addressed by the proposed federated learning pipeline in this work.
\section{Proposed Approach} \label{sec:proposedapproach}
\subsection{Preliminaries}
In this section, we discuss the training of the proposed federated learning framework in subsections. The overall framework is presented in Fig. 1. In the federated learning framework, we assume there are $M$ local clients, where each of them has their local dataset $D_{m}$ containing clean samples $C_{m}$ and adversarial samples $A_{m}$. The distributed paradigm FL aims to learn a central model with the parameter $\theta_c$ over the whole training data $\mathcal{D}=\left\{\mathcal{D}_1, \mathcal{D}_2, \ldots, \mathcal{D}_M\right\}$ using a central model without exchanging local private data. Formally, such a process can be expressed as:
\begin{equation}
\arg \min _{\theta_c} \mathcal{L}(\theta_c)=\sum_{m=1}^M \frac{\left|\mathcal{D}_m\right|}{|\mathcal{D}|} \mathcal{L}_m(\theta_c)
\end{equation}
where the number of samples in $D$ is presented as $|\mathcal{D}|$. $\mathcal{L}_m(\theta_c)$ is the empirical loss of the client $m$ which can be expressed as:
\begin{equation}
\mathcal{L}_m(\theta_c)=\mathbf{E}_{(\mathbf{x}, \mathbf{y}) \in \mathcal{D}^m} \mathcal{L}_m(\mathbf{x} ; \theta_c)
\end{equation}
where $\mathbf{x}$ denotes an image sample with the ground truth $\mathbf{y}$ of dataset $D^{m}$. $\mathcal{L}_m$ denotes the local loss term, \textit{e.g.}, cross-entropy loss.
\begin{figure}[htbp!]
\centering
\includegraphics[width=12.2cm, height=4.4cm]{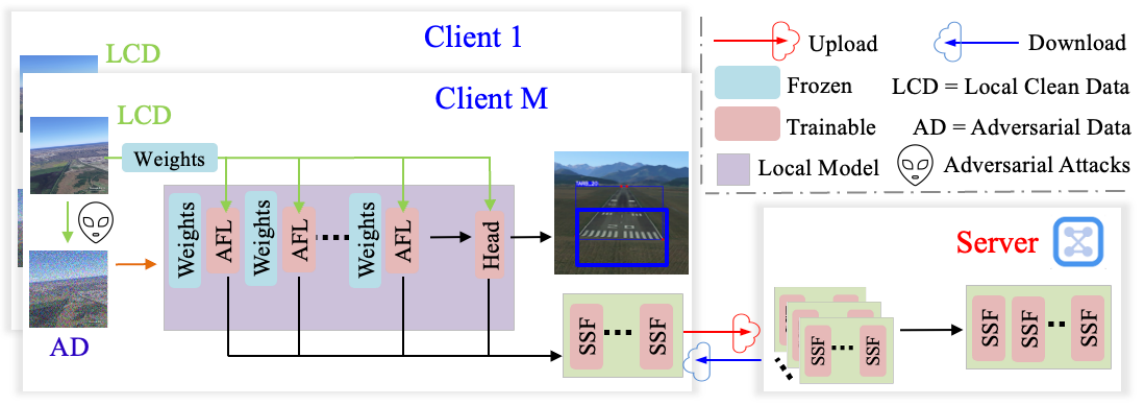}
\caption{Proposed pipeline of the federated learning-based landing runway detection method. The local models are initially pre-trained using lane detection datasets and subsequently fine-tuned with local landing runway detection datasets. The trained Scale and Shift Features (SSF) pools are then aggregated into the final model on the server.}\centering
\label{overallstructure}
\end{figure}
\subsection{Pre-training}
Recent landing runway detection techniques leverage deep learning, but the models in the existing literature are comparatively smaller than state-of-the-art models such as Vision Transformer (ViT) \cite{vit} and YOLO \cite{yolo}. To address this, we propose fine-tuning a large-scale model pre-trained on a similar feature type-based dataset, such as lane detection datasets. This choice is motivated by the similarity between driving lanes and landing runways at the feature level. Specifically, we use a ViT-L/16 model \cite{vit} pre-trained on a road image dataset \cite{tvtLANE} to better extract features from our landing runway dataset. In the diagnostic experiment, detailed analysis of the chosen backbone, ViT-L/16, is provided. 
\subsection{Local Model Fine-tuning}
The pre-trained model is then fine-tuned with airport aerial images from \cite{lard} to learn task-specific knowledge. However, there are two challenges in this stage. Firstly, pre-trained model usually has a large number of model parameters. Direct fine-tuning the full model consequently gives rise to significant communication costs between the server and the client. Secondly, recent studies show that adversarial attacks potentially poison the local data. This causes wrong predictions in autonomous landing systems. Therefore, we aim to mitigate these two challenges by using SSF and adversarial feature learning, respectively.

\subsubsection{Scale and Shift Deep Features}
To efficiently fine-tune each local model, we scale and shift the deep features (SSF) the pre-trained model in the local training phase and merge them into the original pre-trained model weights by reparameterization in the inference phase. Given a pre-trained model with parameter $\theta$, we send the model to each local client and the define the parameter as $\theta_m$ in the $m$-th client. In the fine-tuning stage, the model parameters of SSF can be represented as $\theta_m =\{\gamma_m, \beta_m, h_m, \theta_m\}$, where $\gamma_m \in R^D$ and $\beta_m \in R^D$ are scale and shift factors, respectively \cite{ssf}. $h_m$ is the parameter of the classification head. We break the weights of local models based on operations \cite{ssf}, e.g., multi-head self-attention (MSA), MLP and BN, etc. Then, we remodulate features by inserting SSF with $\gamma_m$ and $\beta_m$ factors after these operations. It is highlighted that the pre-trained weights are kept frozen, and only the SSF and classification head are kept updated. Therefore, we define parameter $\phi_m$ and $\phi_c$ as the combination of trainable $\{\gamma, \beta, h\}$ in the $m$-th local model and central model, respectively. $\phi_c$ can be updated with Eq. (1) as:
\begin{equation}
\phi_c=\arg \min _{\phi \text{s}} \mathcal{L}(\phi_m)=\sum_{m=1}^M \frac{\left|\mathcal{D}_m\right|}{|\mathcal{D}|} \mathcal{L}_m(\phi_m)
\end{equation}
where $\phi$s are the set of SSF, \textit{i.e.}, $\phi_1, \phi_2, ... \phi_m$. In each client, we updates $\phi_m$ in the $r$-th communication rounds between the server and local clients as:
\begin{equation}
\phi_m^{r, e+1} \leftarrow \phi_m^{r, e}-\eta_m \nabla \mathcal{L}_m\left(\mathbf{x}^m, \mathbf{a}^m,; \phi_m^{r, e}\right)
\end{equation}
where $e$ is the index of local updates and $\eta_m$ is the learning rate. Once the local model training is accomplished, $\phi$s can then be merged into $\theta_m$ to obtain the updated model parameter $\theta_m^{\prime}$. Besides, the server performs aggregation every communication round by receiving the updated parameters of local clients after the local updates within each round. Formally, we have
\begin{equation}
\phi_c^{e+1} \leftarrow \sum_{m=1}^M \frac{\left|\mathcal{D}^m\right|}{|\mathcal{D}|} \phi_m^e
\end{equation}
Similarly, once all the communication rounds are accomplished, $\phi_c$s can then be merged into $\theta_c$ to obtain a robust central model parameter without disclosing any local data. 
\subsubsection{Adversarial Feature Learning}
In recent studies \cite{ABN}\cite{IJCNN2024}\cite{UNICAD}, clean and adversarial images have different underlying distributions because the adversarial images essentially involve a two-component mixture distribution. Therefore, in the proposed adversarial feature learning, we aim to disentangle features from the clean and adversarial images to enhance the global feature representation and suppress the adversarial attacks. To achieve that, we generate adversarial images from the clean images by using the attack algorithms, \textit{e.g.}, FGSM. Then, these paired clean and adversarial image samples are fed into the proposed adversarial feature learning (AFL) block. Due to different underlying distributions of clean and adversarial images, different from conventional adversarial image learning techniques \cite{RoCNN}, we exploit different normalization techniques for clean and adversarial images to guarantee its normalization statistics are exclusively preformed on the adversarial images. Particularly, the batch normalization (BN) \cite{BN} and random normalization aggregation (RNA) \cite{RNA} are empirically used for clean and adversarial images, respectively. The support experiments for the chosen experiment configuration will be provided in Section \ref{bn}.


The loss between clean and adversarial images at the $m$-th client is defined:
\begin{equation}
\mathcal{L}_{m} = \frac{1}{N} \sum_{i=1}^{N} \| (\gamma_{\text{cl}} \odot \mathbf{x}_{n}^m+\beta_{\text{cl}} ) - (\gamma_{\text{adv}} \odot \mathbf{a}_{n}^m+\beta_{\text{adv}}\|_2^2
\end{equation}
where $N$ is the number of samples in a local client. The clean and adversarial samples are denoted as $\mathbf{x}_{n}^m$ and $\mathbf{a}_{n}^m$, respectively. Except BN and RNA, clean and adversarial sample parameters $\gamma_{\text{cl}}$, $\beta_{\text{cl}}$, $\gamma_{\text{adv}}$, and $\beta_{\text{adv}}$ for convolutional and other layers are jointly optimized for both adversarial examples and clean images. Specifically, the AFL with an RNA helps to learn the features by keeping separate BNs to features that belong to different domains. 

\subsection{Central Model Update}
When the training is accomplished, we can re-parameterize the SSF by merging it into the original parameter space (\textit{i.e.}, model weight $\theta$). As a result, federated SSF is not only efficient in terms of communication costs, but also does not introduce any extra parameters during the inference phase. The algorithm of the proposed federated adversarial learning framework is presented in Algorithm 1.
\vspace{-0.9em}
\begin{algorithm}
   \caption{ Federated adversarial learning}
   \label{alg:example}
\begin{algorithmic}[1]
   \STATE {\bfseries Input:} Local datasets of $M$ clients: $\mathcal{D}=\left\{\mathcal{D}_1, \mathcal{D}_2, \ldots, \mathcal{D}_M\right\}$, clean samples $\mathbf{x}^m$, adversarial samples $\mathbf{a}^m$, maximum local update $E$, communication rounds $R$, learning rate $\eta_m$, learnable parameters $\phi_m^{r, e}$
   \STATE {\bfseries Output:} The final central model $\theta_c$
   \STATE Initialize SSF 
   \FOR{$e = 1, ...,$  $E$}
   \FOR{$m = 1, ...,$  $M$ in parallel}
   \STATE Download from the central model to local models: $\phi_m^{r, e} \leftarrow \phi_c^{e} $  
   \STATE Local model updates in clients: $\phi_m^{r, e+1} \leftarrow \phi_m^{r, e}-\eta_m \nabla \mathcal{L}_m\left(\mathbf{x}^m, \mathbf{a}^m; \phi_m^{r, e}\right)$
   \ENDFOR
   \STATE Update local models to the central model: $\phi_c^{e+1} \leftarrow \sum_{m=1}^M \frac{\left|\mathcal{D}_m\right|}{|\mathcal{D}|} \phi_m^e$ 
   \STATE  $\theta_c^{e+1}=\phi_c^{e+1}$   
   \ENDFOR
   \STATE SSF Aggregation: $\theta_c=$Agg($\theta_c^{1}, \theta_c^{2}, ..., \theta_c^{E}$)
\end{algorithmic}
\end{algorithm}
\vspace{-0.9em}
\section{Experimental Results} \label{sec:results}
\subsection{Dataset and Attacks}
The tvtLANE dataset \cite{tvtLANE} contains 19,383 image sequences for lane detection, and 39,460 frames of them are labeled. In this work, the model is pre-trained with randomly selected 35,000 images from the dataset. We further fine-tune the local models on synthetic and real runways from the LARD dataset \cite{lard}. The LARD dataset contains 14,433 training images of resolution 2448$\times$2648, taken from 32 runways in 16 different airports in total. In the training stage, we set number of clients to 5, and randomly select 2,800 for each client. Then, we construct the validation set with the rest 433 images. In the test stage, we evaluate the central server with 2,221 synthetic images taken from 79 runways in 40 different airports and 103 hand-labeled pictures from real landing footage on 38 runways in 36 different airports.

We select aforementioned attacks in Section II because they are robust to novel adversarial attack recovery techniques \cite{IJCNN}\cite{detecting}. The adversarial images in the training and test stages are generated with the same attack algorithm.



\subsection{Competitors and Performance Measure}
In this paper, the proposed method is evaluated and compared to state-of-the-art competitor models. Firstly, we select four landing runway detection techniques, including long short-term memory (LSTM) \cite{lrd4}, Line Segment Detector (LSD) \cite{lrd1}, Runway Detection Systems (RDS) \cite{lrd2}, Complex Cross-Residual Network (CS-ResNet) \cite{lrd3} as the original implementations in the literature but with same data as the proposed method. Secondly, we use four federated learning frameworks, including Siloed Batch Normalization (SiloBN) + Adaptive Sharpness-Aware Minimization (ASAM) \cite{asam}, federated optimization (FedProx) \cite{flprox}, federated averaging (FedAvg) \cite{fl0}, and federated local drift decoupling and correction (FedDC) \cite{flddc} which are state-of-the-art in image processing tasks to confirm the proposed model is robust to adversarial attacks. For a fair comparison, we reproduce these models with same data as the proposed method. We calculate the average error between 6 predictions and ground truths (e.g., along track distance, vertical path angle, lateral path angle, yaw, pitch, and roll).

\subsection{Model Configuration}
We pre-train a Vit-L/16 model \cite{vit} as the backbone by using PaddleSeg toolkit \cite{roadgit}. The backbone study is presented later in diagnostic experiments. The model is trained by using the SGD optimizer with a weight decay of 0.0001, a momentum of 0.9, and a batch size of 256. We train the model for 300 epochs. The initial learning rate is 0.03, and is multiplied by 0.1 at 500 and 1000 epochs.

After the pre-training, we only fine-tune the SSF layers and freeze the other weights of the Vit-L/16 model. In particular, we fine-tune the model for 100 epochs. All experiments are run on the High End Computing (HEC) Cluster with Tesla V100 GPUs.
\subsection{Comparison to State-of-the-Arts} \label{sota}
We conduct two experiments in this section. In the first experiment, we assume the federated learning perfectly protect the data privacy. Therefore, the central model is evaluated with \textit{clean} samples of LARD \cite{lard}. Table 1 shows the results, each of them is the average of 2,221 synthesis images or 103 real images. In the second experiment, we evaluate the performance over \textit{adversarial} data and present the results in Fig. 2(2).
\begin{table}[htbp!]
\centering
\caption{Detection error comparison with ALRD and federated learning methods on the \textbf{clean} samples of LARD dataset. Para. denotes the \textbf{trainable} parameters.}
\begin{tabular}{c|cc|cc}
\hline
Method & FL &  Para. (M) & Synthesis ($\%$) & Real ($\%$) \\ 
\hline
LSD \cite{lrd1} & \xmark & 25.6  & 28.6 & 37.5\\ 
LSTM \cite{lrd4} &  \xmark & 0.5 & 28.0 & 35.8\\
RDS \cite{lrd2} &  \xmark & 38.9 & 25.2 & 34.1\\
CS-ResNet \cite{lrd3} & \xmark & 58.2 & 22.8 & 31.9\\
\hline
FedAvg \cite{fl0} & \checkmark & 1.7 & 29.5 & 37.6\\
FedProx \cite{flprox} & \checkmark & 53.2 & 26.1 & 35.2\\
FedDC \cite{flddc} & \checkmark & 11.2 & 25.0 & 35.0\\
SiloBN + ASAM \cite{asam} & \checkmark & 4.5 & 24.8 & 32.6\\
\hline
\textit{Ours} &\checkmark  & 7.4 & {\bfseries 19.5} & {\bfseries 26.3} \\
\hline 
\end{tabular}
\end{table}

Table 1 shows that our federated adversarial learning improves results over landing runway detection models and federated algorithms. Our model is 3.3 points lower than CS-ResNet (19.5$\%$ vs. 22.8$\%$). We conduct more experiments to confirm this point in Fig. 2. Adversarial images are generated from clean samples of LARD \cite{lard} for the model training and evaluation. Fig.2 (a) shows the results, each of them is the average of 15,547 (2,221 synthesis samples) synthesis images or 721 (103 real samples $\times$ 7 attack algorithms) real images.
\begin{figure}[ht]
  \centering
   \includegraphics[width=12.2cm, height=4.2cm]{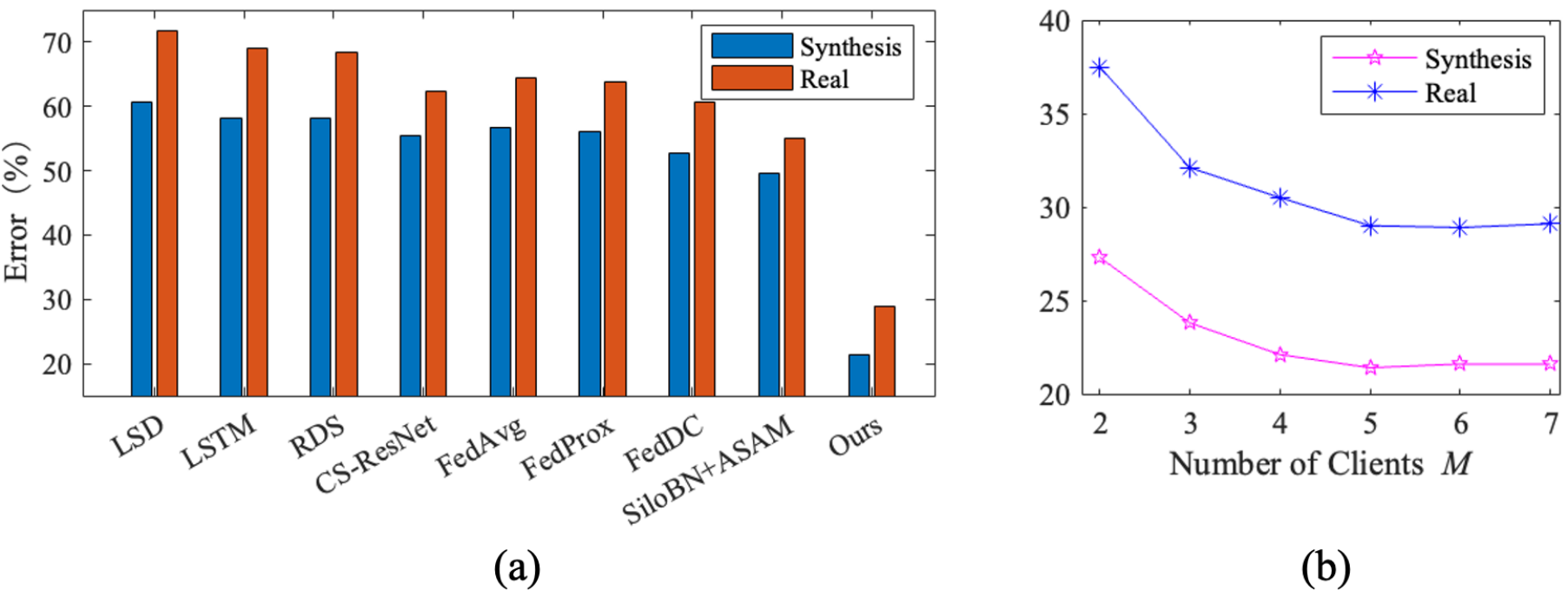}
   \caption{Detection error comparison (a) to competitor models (b) against number of clients over LARD.}
   \label{fig:axial}
\end{figure}

From Fig.2(a), it can be observed that in all the evaluated models, the proposed model achieves 21.4$\%$ and 29.0$\%$ for synthesis and real images detection, respectively, which offers the best effectiveness. These observations are consistent with clean images. Moreover, comparing the results to Table 1, competitor models suffer a significant accuracy degradation with adversarial attacks, while the proposed models perform more robust because the global feature representation is enhanced to suppress the adversarial attacks.
\subsection{Diagnostic Experiment}
In this section, we first conduct experiments to validate several intriguing properties, \textit{e.g.}, backbone. Then, we study the efficacy of our core ideas and essential pipeline design.
\subsubsection{Number of Clients}
We conducted experiments to demonstrate the trade-off between performance improvement and network depth, specifically, varying the number of proposed PAA blocks. Fig. 2(b) presents these results, with each data point being an average of 15,547 (2,221 synthesis samples $\times$ 7 attack algorithms) synthesis images or 721 (103 real samples $\times$ 7 attack algorithms) real images.


Fig. 2(b) compares detection error against the number of clients on LARD. As Fig. 4 shows, detection errors start to decrease from $M=2$, but performance is fairly stable for 5 $\le M \le$ 7. Therefore, the results indicate that $M=5$ offers the best trade-off, validating the chosen implementation setting.
\subsubsection{Backbone}
We implement the proposed federated learning framework with different backbones. The backbones are pre-trained on the tvtLANE dataset \cite{tvtLANE} and fine-tuned in clients. The experimental results are provided in Table 2. Each result of them is the average of 15,547 (2,221 synthesis samples $\times$ 7 attack algorithms) synthesis images or 721 (103 real samples $\times$ 7 attack algorithms) real images.

\begin{table}[htbp]
    \begin{minipage}{.5\linewidth}
        \centering
        \caption{Backbones.}
\begin{tabular}{c|cc|cc}
\hline
 & \multicolumn{2}{c|}{Para. (M)}  & \multicolumn{2}{c}{Error ($\%$)} \\
 \hline
 & Backbone & SSF & Synthesis & Real \\
 \hline
ResNet50 \cite{rc} & 25.6 & - & 28.5 & 36.8 \\
ResNet101 \cite{rc} & 42.8 & - & 28.1 & 35.6 \\
ENet-B3 \cite{det} & 12.0 & - & 25.8 & 39.1 \\
ENet-B5 \cite{det} & 30.6 & - & 24.7 & 38.0 \\
ENet-B7 \cite{det} & 66.0 & - & 24.0 & 36.3 \\
WRN-50-2 \cite{WRN} & 68.9 & - & 29.7 & 38.2 \\
WRN-101-2 \cite{WRN} & 126.8 & - & 28.0 & 35.6 \\
VGG16 \cite{VGG} & 138.3 & - & 31.4 & 42.9 \\
ViT-S-16 & 22.0 & 3.96 &  26.6 & 35.8\\
ViT-B-16 & 86.5 & 4.62 &  23.0 & 33.1  \\
ViT-L-16 & 307.1 & 7.39 & {\bfseries 21.4} & {\bfseries 29.0}\\
 \hline 
\end{tabular}
        
    \end{minipage}%
    \begin{minipage}{.6\linewidth}
        \centering
        \caption{Normalization techniques.}
\begin{tabular}{c|c|ccccc}
 \hline
                      &    & \multicolumn{5}{c}{Adversarial}    \\
 \hline
                      &    & BN & LN & IN & GN & RNA\\
 \hline
\multirow{5}{*}{Clean} & BN & 27.5  & 31.9 & 38.2 & 36.8 &{\bfseries 23.4}\\
                      &  LN & 32.6 & 34.7 & 40.6 & 31.2 & 33.3 \\
                      & IN  & 35.9 &  41.2 &  40.5 & 30.7 & 29.8 \\
                      &  GN & 30.4 & 29.8 & 35.7 & 29.1  & 31.8 \\
                      & RNA & 25.8 & 29.5 & 34.2 & 33.9 & 25.9 \\
 \hline
\end{tabular}
        
    \end{minipage}
\end{table}


Table 2 compares the detection error against backbone networks on LARD. The results indicate that: 1) Although the ViT family requires massive network parameters, the SSF significantly reduces the parameters, which facilitates the training; 2) ViT-L-16 with SSF offers the best trade-off between the detection performance and computational cost, supporting the chosen experiment configuration. 
\subsubsection{Normalization Techniques} \label{bn}
A diagnostic experiment of normalization techniques including BN \cite{BN}, layer normalization (LN) \cite{LN}, instance normalization (IN) \cite{IN}, group normalization \cite{GN}, and RNA \cite{RNA} for clean and adversarial images is conducted on the LARD dataset. Each result in Table 3 is an average of 16,268 experiments (2,221 synthesis samples $\times$ 7 attack algorithms + 103 real samples $\times$ 7 attack algorithms).

According to Table 3, BN and RNA are optimal for clean and adversarial image features, respectively. The detection error achieves 23.4$\%$ at the valley, which supports the experiment configuration. 

\subsubsection{Ablation Study}
In this section, we investigate the effectiveness of each contribution based on the LARD dataset. The ablation study is presented in Table 4 and the experimental setting is the same as Section \ref{sota}. Each result is the average of 15,547 (2,221 synthesis samples $\times$ 7 attack algorithms) synthesis images or 721 (103 real samples $\times$ 7 attack algorithms) real images.

\begin{table}[htbp!]
\caption{Ablation study in the proposed method. afl and fl denote adversarial feature learning and federated learning, respectively.}
\centering
\begin{tabular}{cccccc}
\hline
\multicolumn{3}{c}{Ablation Settings} & %
    \multirow{2}{*}{Para. (M)} & %
    \multirow{2}{*}{Synthesis ($\%$)}  & \multirow{2}{*}{Real ($\%$)} \\
\cline{1-3}
  AFL & FL & SSF \\
 \hline
\xmark &\xmark &\xmark & 307.1 & 52.5 & 59.1 \\
\checkmark   &\xmark &\xmark   & 307.1 & 31.6 & 38.3\\
\xmark  &  \checkmark &\xmark  & 1535.5 & 44.2 & 50.7 \\
\xmark  &  \xmark &\checkmark  & 1.5 & 52.3 & 59.0 \\
 \hline 
\xmark  &  \checkmark &\checkmark  & 7.39 & 43.9 & 48.5\\
\checkmark   &\xmark  &\checkmark & 1.5 & 34.8 & 42.6 \\
\checkmark   &\checkmark  &\xmark &  1535.5 & 21.5 & 29.5 \\
 \hline 
\checkmark   &\checkmark   &\checkmark &7.39 & {\bfseries 21.4 }& {\bfseries 29.0} \\
 \hline 
\end{tabular}
\end{table} 

Initially, we evaluate the effectiveness of AFL, which plays a pivotal role in learning desired features from adversarial images. AFL demonstrates its significant impact within the federated learning framework and SSF, resulting in a remarkable performance improvement from an initial error rate of 43.9$\%$ to 21.4$\%$. This improvement can be attributed to the enhanced global feature representation provided by AFL, effectively suppressing adversarial attacks.

Moreover, the detection error experiences a significant reduction by exploiting federated learning framework (\textit{i.e.}, 34.8$\%\rightarrow$ 21.4$\%$, synthesis images). The SSF aggregation captures diverse features learned from different local clients and data, empowering the central client to make more accurate predictions through the assimilation of rich features.

The final experiment in the ablation study involves the addition of SSF. Consequently, federated SSF not only prove to be efficient in terms of communication costs but also do not introduce any extra parameters during the test stage.
\subsection{Visualizations}
In this section, we present qualitative results demonstrating the landing runway detection of attacked image samples on LARD \cite{lard} in Fig. 3.

\begin{figure}[ht]
  \centering
   \includegraphics[width=8.8cm, height=7cm]{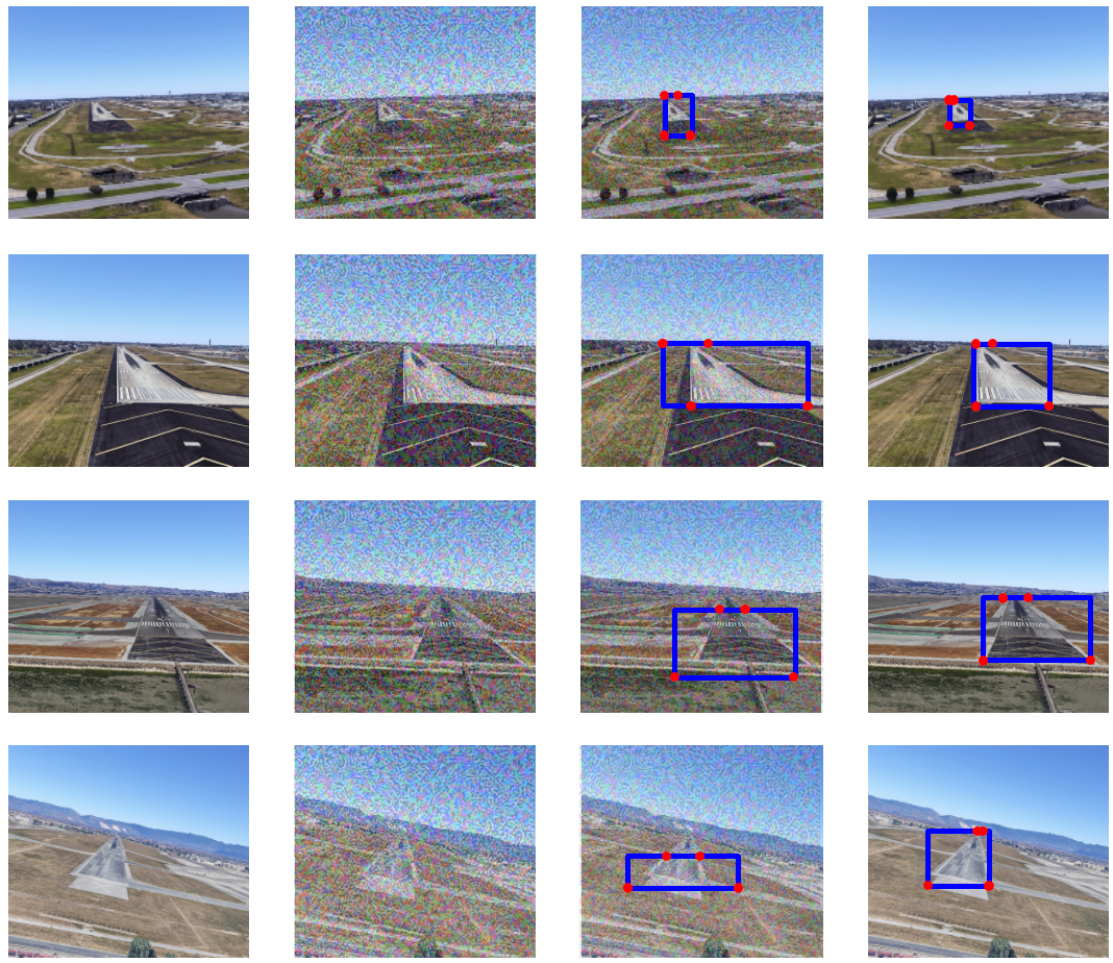}
   \caption{Qualitative landing runway detection results on LARD. From left to right: original images, images attacked by FGSM, results of the central model without adversarial training, results of the central model.}
   \label{fig:axial}
\end{figure}

After comparing the reconstructed images with the original and attacked images, it can be observed that the detection boxes obtained via the proposed model with adversarial training provide more accurate descriptions of the landing runway areas. This observation further confirms the efficacy of the proposed method.


\section{Conclusion} \label{sec:conlusion}
In this paper, we have proposed a federated adversarial learning framework as a simple yet effective alternative to conventional landing runway detection algorithms. To efficiently fine-tune the pre-trained local model on clients, we utilize a technique of shifting and scaling features with both clean and adversarial samples. Subsequently, the SSF pools are aggregated into the central model. Our evaluation on LARD has demonstrated the high efficiency and effectiveness of the proposed model through the use of qualitative results and quantitative results which includes detection error comparison between ALRD and FL, different backbones, different normalisation techniques and lastly a thorough ablation study amongst others.
\section*{Acknowledgment}
Research supported by the UKRI Trustworthy Autonomous Systems Node in Security/EPSRC Grant EP/V026763/1.
\bibliographystyle{IEEEtran}
\bibliography{samplepaper}

\end{document}